\documentclass[preprint,12pt]{elsarticle}

\usepackage{amssymb}
\usepackage{subfig}
\usepackage{algorithmic}
\usepackage{algorithm}
\usepackage{amsmath}
\usepackage{multirow}

\journal{Journal of biomedical informatics}

\begin{document}

\begin{frontmatter}



\title{Review learning: Real world validation of privacy preserving continual learning across medical institutions}


\author[KU]{Jaesung Yoo}
\author[SNUHBE]{Sunghyuk Choi}
\author[Uijeongbu]{Ye Seul Yang}
\author[CAU]{Suhyeon Kim}
\author[CAU]{Jieun Choi}
\author[CAU]{Dongkyeong Lim}
\author[CAU]{Yaeji Lim}
\author[KUMC]{Hyung Joon Joo}
\author[AJOUMC]{Dae Jung Kim}
\author[AJOUMCBI]{Rae Woong Park}
\author[SNUHBE]{Hyung-Jin Yoon}
\author[Uijeongbu,SNUHT]{Kwangsoo Kim}

\affiliation[KU]{organization={School of Electrical Engineering, Korea University}}
\affiliation[SNUHBE]{organization={Department of Biomedical Engineering, Seoul National University College of Medicine}}
\affiliation[Uijeongbu]{organization={Department of Medicine, Seoul National University College of Medicine}}
\affiliation[CAU]{organization={Department of Applied Statistics, Chung-Ang University}}
\affiliation[KUMC]{organization={Department of Cardiology, Cardiovascular Center, College of Medicine, Korea University}}
\affiliation[AJOUMC]{organization={Department of Endocrinology and Metabolism, School of Medicine, Ajou University}}
\affiliation[AJOUMCBI]{organization={Department of Biomedical Informatics, School of Medicine, Ajou University}}
\affiliation[SNUHT]{organization={Transdisciplinary Department of Medicine \& Advanced Technology, Seoul National University Hospital}}

\begin{abstract}
When a deep learning model is trained sequentially on different datasets, it often forgets the knowledge learned from previous data, a problem known as catastrophic forgetting. This damages the model's performance on diverse datasets, which is critical in privacy-preserving deep learning (PPDL) applications based on transfer learning (TL). To overcome this, we introduce “review learning” (RevL), a low cost continual learning algorithm for diagnosis prediction using electronic health records (EHR) within a PPDL framework. RevL generates data samples from the model which are used to review knowledge from previous datasets. Six simulated institutional experiments and one real-world experiment involving three medical institutions were conducted to validate RevL, using three binary classification EHR data. In the real-world experiment with data from 106,508 patients, the mean global area under the receiver operating curve was 0.710 for RevL and 0.655 for TL. These results demonstrate RevL's ability to retain previously learned knowledge and its effectiveness in real-world PPDL scenarios. Our work establishes a realistic pipeline for PPDL research based on model transfers across institutions and highlights the practicality of continual learning in real-world medical settings using private EHR data.

\end{abstract}

\begin{graphicalabstract}
\includegraphics[width=\textwidth]{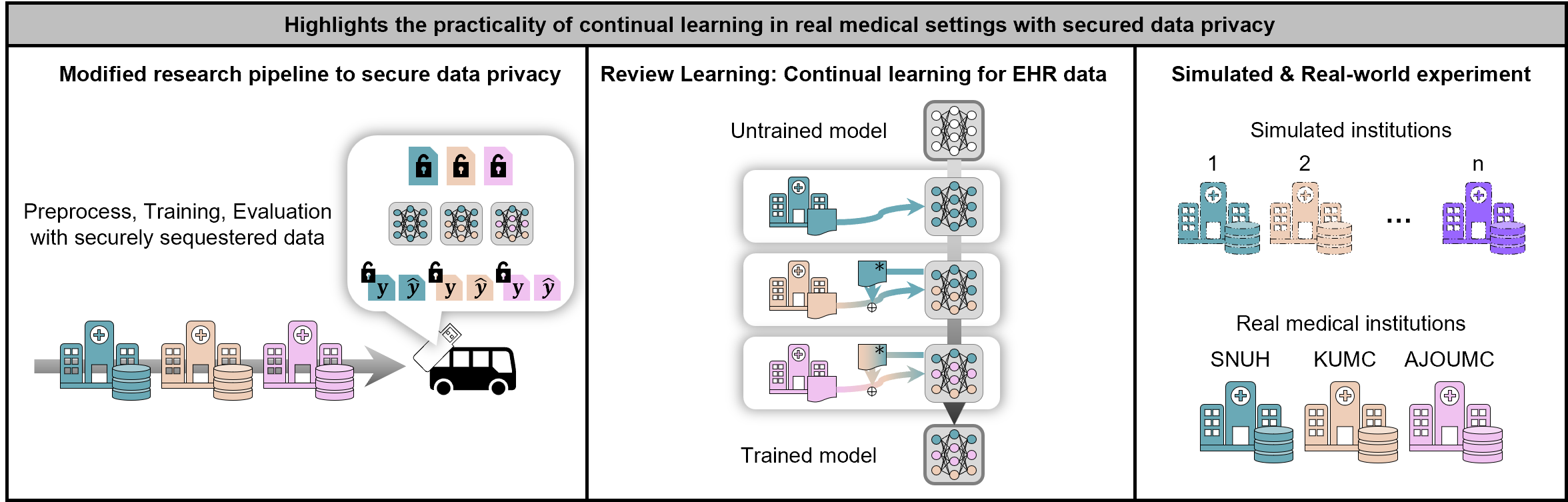}
\end{graphicalabstract}

\begin{highlights}
\item End to end research pipeline was modified to secure data privacy.
\item A low cost continual learning algorithm for EHR data was proposed.
\item Real-world experiment with three medical institutions was performed.
\item Results show the practicality of continual learning in real medical PPDL settings.

\end{highlights}


\end{frontmatter}



\section{Introduction}
Deep learning (DL) \cite{RN38} has empirically achieved outstanding performance in various domains when abundant, high-quality data is available \cite{RN36,RN23,RN106,RN202, YOO2022107079, YU2024124551, YU2024105223, yoo2023deep}, despite the lack of convergence proofs due to its non-convex nature \cite{NEURIPS201890365351, doi:10.1137/030601296, pmlr-v97-allen-zhu19a}. However, data are often limited due to natural scarcity from constrained data collection bandwidth \cite{RN66, dash2019big, tambe2023integrative} or data privacy issues \cite{RN69,RN61, kim2024continual}, which are often encountered in biomedical studies \cite{dash2019big, kim2024continual, yang2021intelligent, RN203, RN184}. Among different approaches to preserve data-privacy while developing a DL model \cite{jonnagaddala2025privacy} (e.g. privacy-preserving data publishing \cite{kara2023collection, 9255155, kara2023new, 10.1007/978-3-031-09753-9_12}), transfer learning (TL) \cite{RN93,RN11, 9134370} is a training algorithm based privacy-preserving deep learning (PPDL) \cite{RN56,RN77} method to leverage data from different institutions, by training the model sequentially through institutional transfers. TL is an effective PPDL method that requires minimal asynchronous institutional communication, since other PPDL methods such as federated learning \cite{RN187, RN63, RN49, RN50, WU2024108905} require heavy synchronization from automated platforms \cite{RN51, RN55} which may not be available in medical institutions with strict security protocols \cite{RN99} or limited technical infrastructure \cite{RN98}.

Despite its popular use \cite{RN54,RN94,RN52,RN62,RN75,RN64, ZHAO2024108348, DARVISHIBAYAZI2024107893, PHAM2025109461}, TL has a weakness: the catastrophic forgetting problem \cite{RN25,RN196,goodfellow2013empirical,10341211,RN95}. Unlike the biological brain \cite{RN198,RN199,RN200,RN201}, which serves as both the origin and recurring inspiration for improving DL models \cite{hassabis2017neuroscience, richards2019deep, NIPS2013_8f1d4362, yoo2024dual}, the DL model forgets knowledge from previous data when sequentially trained on data from different institutions. Catastrophic forgetting is a challenge in PPDL applications where the goal is to learn valuable knowledge from privacy-sensitive data distributed across institutions without sharing them \cite{RN52}.

To address this, several studies on continual learning \cite{10341211, RN95, RN204, FERRI2024108548, SHI2024109028, AMMOUR2021104807} have focused on training DL models to acquire diverse knowledge from sequential training. Continual learning is typically categorized into settings such as task-incremental learning, domain-incremental learning, and class-incremental learning, depending on the input-output relationships the DL model must learn \cite{van2022three}. Various methods have been proposed for continual learning \cite{10341211, RN95}, including regularization approaches \cite{RN26, 8107520}, dynamic architectures \cite{rusu2016progressive, pmlr-v70-cortes17a}, and memory replay \cite{RN192, 7965898}, each of which remains an active area of research. In this study, we focus on the memory replay based continual learning method.

The DL model may be continually trained by storing previous data into a storage and replaying them when training with the next institutional data. Although this direct replay is promising \cite{RN197}, accumulating previous data is not scalable \cite{RN95,schwarz2018progress} and highly restricted in PPDL settings. As an alternative, a generative model is built for the main DL model to access a simile of the previously trained data when in a new institution \cite{RN192}. This continual learning method, known as generative replay, has an improved continual learning performance compared to the standard TL \cite{RN96,RN205}. Nonetheless, it is more computational workload \cite{RN194,RN195} as a generative model must be additionally trained along with the main task model, which often requires difficult fine tuning to prevent mode collapse \cite{saad2024early, 9312049, 9207181, Bau_2019_ICCV}. Furthermore, the generative model is an additional burden to be carried across institutions in PPDL setting. To address this limitation, a replay-based continual learning approach without a generator was proposed using image data \cite{pourkeshavarzi2021looking}. However, its effectiveness on tabular data, such as electronic health records (EHR) \cite{BERNARDINI2019103358, SCHILCHER2024107704, WENG2024107687}, remains uncertain due to the differences in data modality. Also, implementing this approach in real-world medical PPDL settings requires more than just the continual learning algorithm; it also necessitates significant research pipeline modifications. Furthermore, the approach's real-world effectiveness remains unverified, as it was only tested on open-source benchmark data.

In this study, we address such restrictions of current medical PPDL settings; The necessity of using TL-based methods for PPDL, leveraging continual learning to alleviate the forgetting problem in TL, the computational benefits of using generator-free replay-based continual learning; and its unexplored application to EHR data within a PPDL pipeline and real-world settings.

This study makes the following contributions. First, we make research pipeline adjustments to realistically implement PPDL with EHR data across institutions. Second, we introduce “review learning” (RevL), a generator-free continual learning algorithm that uses feature visualization \cite{RN155} to generate data within the main EHR classification model, removing the need for a separate generator and reducing both training complexity and institutional transfer overhead. RevL was designed from the intuition of extracting meaningful features from the learned DL model and reviewing them to learn continually. Third, we compare RevL with the baseline PPDL model development methods used in medical practice, namely TL, collaborative data sharing (CDS), and local learning (LL) \cite{RN52}. We validated RevL's generalizable performance through PPDL experiments on six simulated institutional settings with feature heterogeneity using two distinct EHR data. Next, we further validated its performance on real-world medical institutions using EHR data that were securely sequestered at remote institutions including Seoul National University Hospital (SNUH), Korea University Medical Center (KUMC), and Ajou Medical Center (AJOUMC). The DL models trained with RevL showed improved prediction performance compared to models trained with TL, on datasets from previous institutions in both simulated and real-world medical institutions. Therefore, our work demonstrates that RevL is a generator-free continual learning algorithm that is effective in real-world PPDL settings using EHR data and highlights the applicability of continual learning in medical PPDL setting.

Section 2 outlines the methods of our study, including the adjusted end-to-end PPDL research pipeline, preprocessing, review learning, evaluation, and experimental details. Section 3 describes the data materials used in both simulated and real-world institutional experiments. Section 4 presents the results from all six simulated institutional experiments and the real-world institutional experiment. Section 5 discusses the findings, highlights key aspects of the review process in review learning, and proposes future research directions to address the study's limitations. Finally, Section 6 provides the conclusion of the study.

\section{Methods}

\begin{figure*}[!t]
\centering
\subfloat[]{\includegraphics[width=0.345\textwidth]{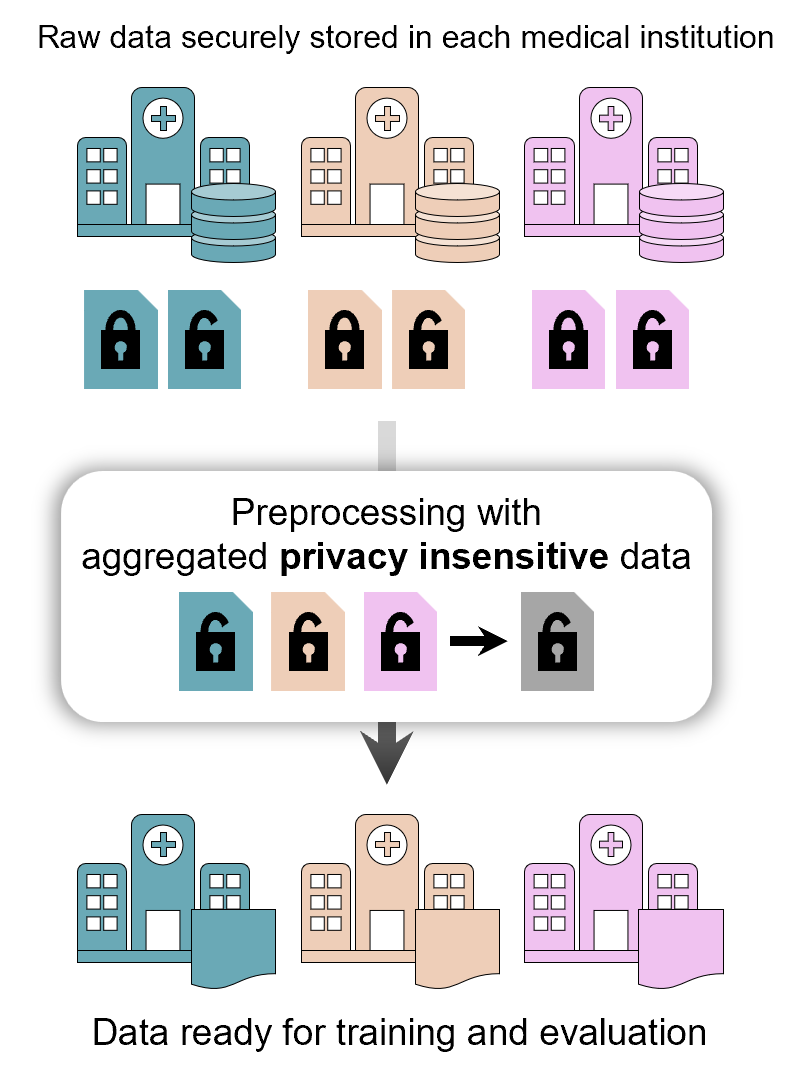}%
\label{fig:data_preparation}}
\hfil
\subfloat[]{\includegraphics[width=0.30\textwidth]{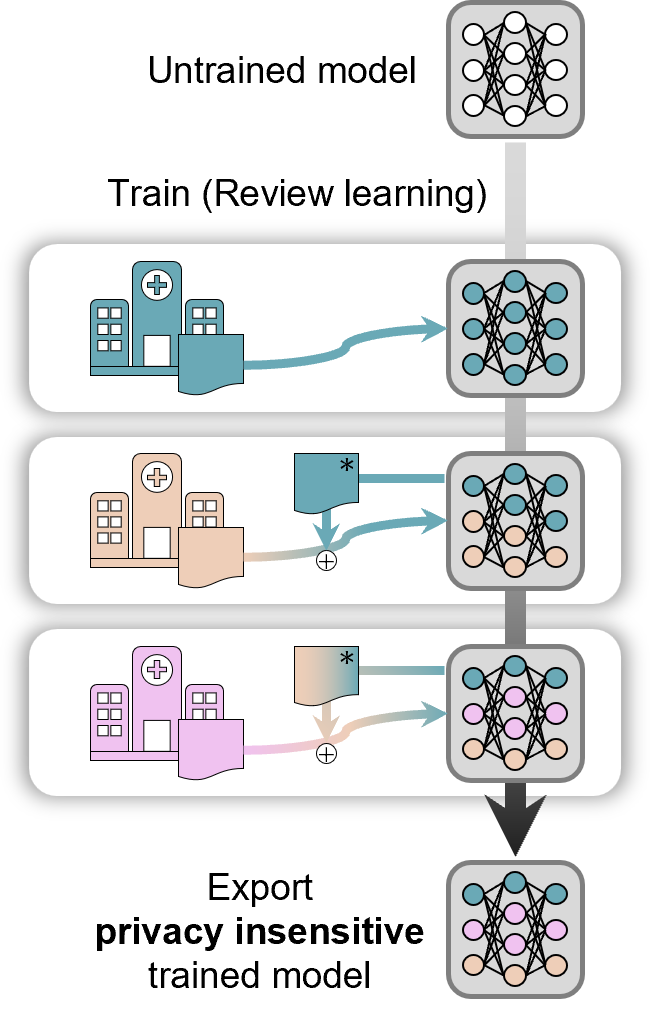}%
\label{fig:model_training}}
\hfil
\subfloat[]{\includegraphics[width=0.345\textwidth]{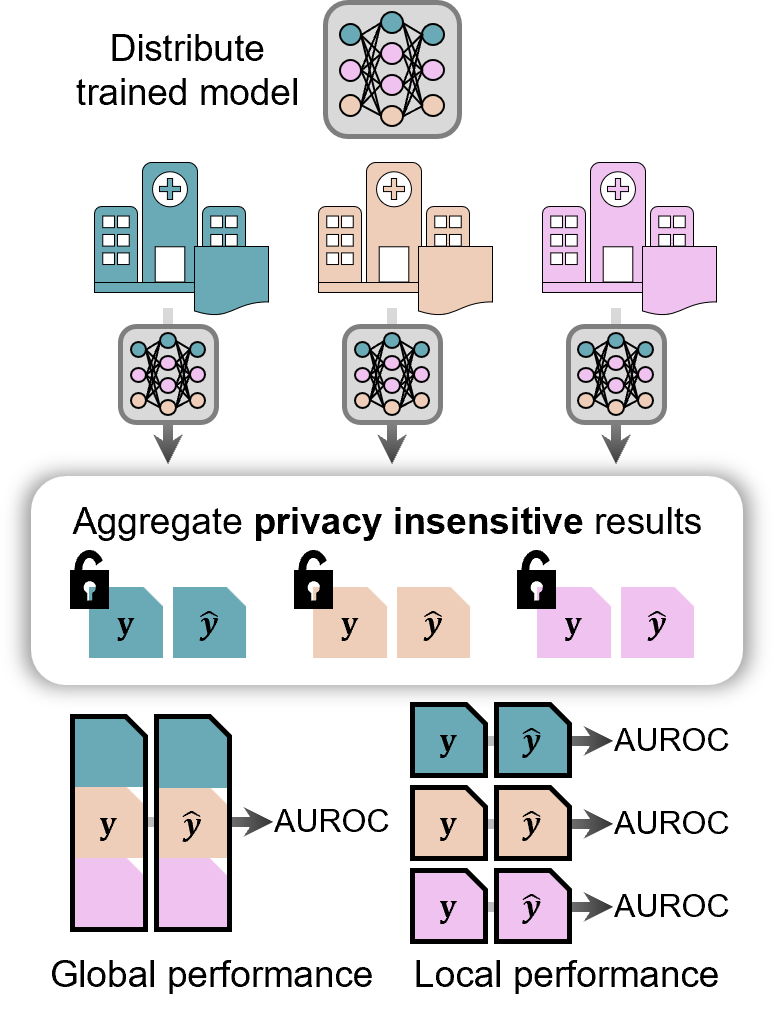}%
\label{fig:model_evaluation}}
\caption{Overview of privacy-preserving deep learning (PPDL) research pipeline. (a) Data preprocessing stage. (b) Model training stage. (c) Model evaluation stage. Abbreviations: area under the receiver operating characteristic curve (AUROC)}
\label{fig:privacy_preserving_deep_learning}
\end{figure*}

\subsection{Privacy Preserving Deep Learning Research Pipeline}
The DL research pipeline was divided into three stages that needed modifications to avoid aggregating data and only share privacy insensitive files (Fig. \ref{fig:privacy_preserving_deep_learning}). In the data preparation stage, the data stored in each institution were preprocessed by sharing the privacy-insensitive statistics (metadata). In the model training stage, the models were trained using training algorithms, including RevL and TL. The privacy insensitive model parameters were exported after training in each institution. In the model evaluation stage, trained models were copied to each institution, where the models predicted test data. The evaluation metrics were measured in global and local scores, which demonstrated the prediction performance on data from all and each institution, respectively. The global score was measured by aggregating privacy-insensitive prediction results for the test data.

\subsubsection{Preprocessing}
\label{sec:preprocess}
Patients from each EHR dataset contained binary, categorical, and continuous variables, and a single feature vector was generated for each patient. Binary variables had a value of 1 if there was a record, and 0 otherwise. Categorical variables were converted into multiple binary columns and encoded in the same way. Continuous variables were standardized using the statistics from the training set. For continuous variables that do not exist for a patient, they were imputed with 0 value after preprocessing, which corresponds to having a mean value of that variable. A value of 0 was set to minimize the missing variable's impact on updating the DL model's weights, as this results in zero gradients. If a new institution introduces additional continuous variables, the model can add new columns without affecting its predictions for existing institutions, as the input for those variables will be set to 0. The final input to the DL model was a vector with standardized continuous variables and binary variables with values 0 or 1.

Due to data privacy constraints, direct data aggregation was impossible. Therefore, global statistics were indirectly derived from local statistics using equations (\ref{eq:global_N})–(\ref{eq:global_std}).

\begin{equation}
\label{eq:global_N}
N=\sum_{i}^{k} N_i
\end{equation}
\begin{equation}
\label{eq:global_mean}
\mu=\frac{\sum_{i}^{k} N_{i}\mu_{i}}{\sum_{i}^{k} N_{i}}
\end{equation}
\begin{equation}
\begin{aligned}
\label{eq:global_std}
\sigma=&\frac{\sum_{i}^{k}N_{i}(\mu_i^2+\sigma_i^2)}{N} - \mu^2 \\
=&\frac{\sum_{i}^{k}N_{i}(\mu_i^2+(\frac{1}{N_i}\sum_{j}^{N_i} x_{i,j}^2-\mu_i^2))}{N} - \mu^2\\
=&\frac{\sum_{i}^{k} \sum_{j}^{N_i} x_{i,j}^2}{N} - \mu^2\\
=&E(X^2)-E(X)^2
\end{aligned}
\end{equation}
where $k$ is the number of institutions, $i$ the institution number, $x_{i,j}$ the $j$th data from institution $i$, $\mu$ the mean value, $\sigma$ the standard deviation, and $N$ the number of samples. Note that $\mu$, $\sigma$, and $N$ refer to the global statistics, whereas $\mu_i$, $\sigma_i$, and $N_i$ refer to the local statistics from institution $i$.

\subsubsection{Review Learning}

\begin{figure*}[!t]
\centering
\includegraphics[width=\textwidth]{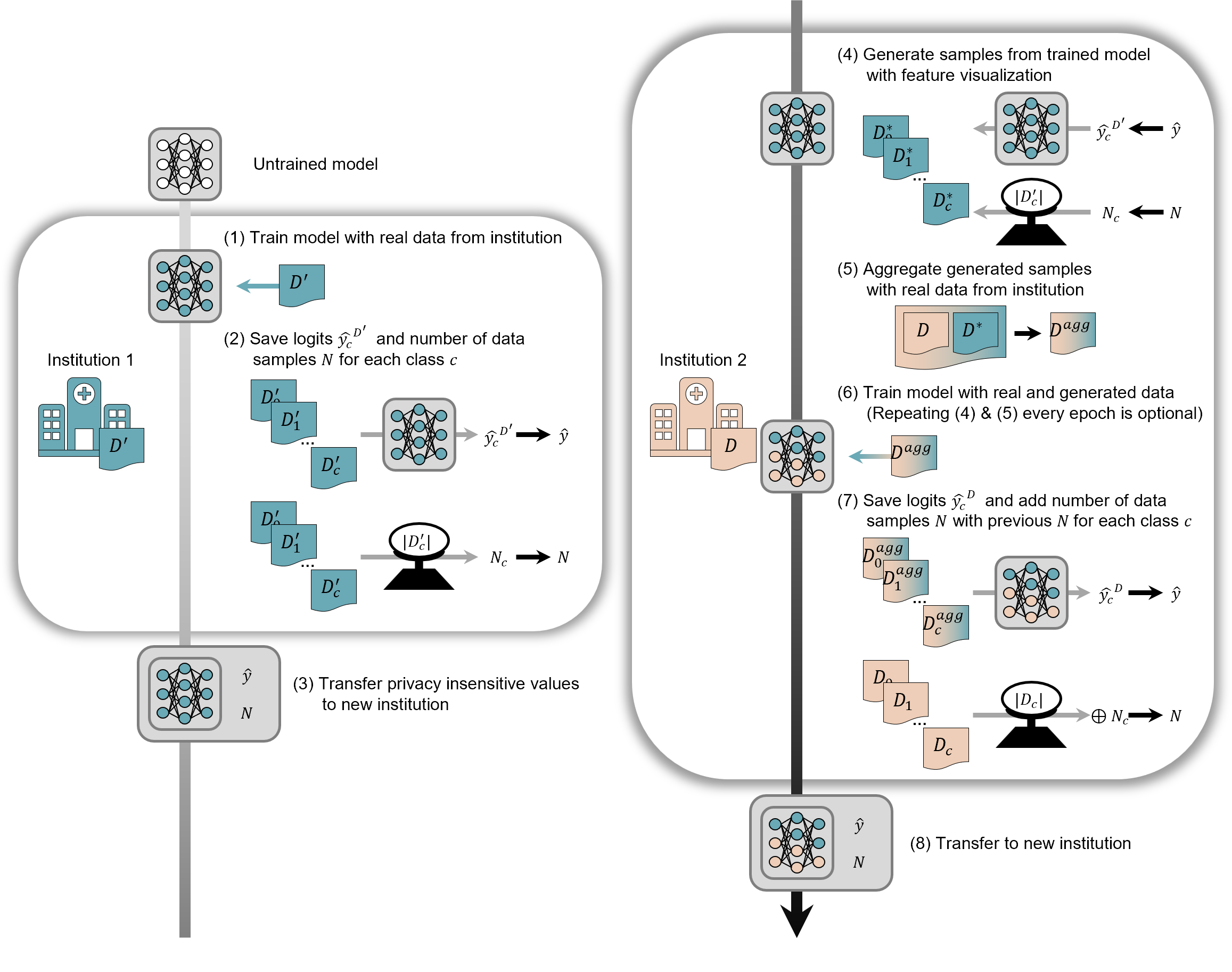}
\caption{Review Learning Diagram}
\label{fig:review_learning}
\end{figure*}

RevL is a new continual learning algorithm based on generative replay. The algorithm is named after its reviewing process to retain previous knowledge by generating representative samples from the model and training the model with them. The RevL algorithm is presented in Algorithm \ref{alg:review_learning} and Fig. \ref{fig:review_learning}.

\begin{algorithm}[ht]
\caption{Review Learning.}\label{alg:review_learning}
\begin{algorithmic}
\STATE $\theta_i$: Model parameters after training from institution $i-1$
\STATE $D$: Current dataset
\STATE $D^{'}$: Previous dataset
\STATE $\hat{y}_{c}^{D^{'}}$: Logits of previous dataset
\STATE $N_{review}$: Number of data samples the model was trained on
\STATE $N_{real}$: Number of data samples in $D$
\STATE
\STATE $\theta_i^* \gets \theta_i$ $//$ Copy model for knowledge extraction
\STATE \textbf{for} epoch \textbf{in} max\_epoch \textbf{do}
\STATE \hspace{0.5cm} $//$ Generate data with feature visualization
\STATE \hspace{0.5cm} $D^*, \hat{y}^*_c \gets$ KnowledgeExtraction($\theta^*_i, \hat{y}_c^{D^{'}}$)
\STATE \hspace{0.5cm} $//$ Jointly optimize with current and generated data
\STATE \hspace{0.5cm} $\theta_i \gets$ KnowledgeDistillation($\theta_i, D, D^*, \hat{y}_c^*$)
\STATE \textbf{end}
\STATE $\theta_{i+1} \gets \theta_i$
\STATE $//$ New review learning parameters
\STATE $\hat{y}_c^D \gets$ MeasureLogits($\theta_{i+1}, D, D^*$)
\STATE $N_{review} \gets N_{review} + N_{real}$
\STATE
\STATE \textbf{return} $\theta_{i+1}, \hat{y}_c^D, N_{review}$
\end{algorithmic}
\label{alg1}
\end{algorithm}

RevL is composed of two steps: knowledge extraction and knowledge distillation. In the knowledge extraction step, data samples are extracted from the model by applying feature visualization \cite{RN155}, described below. First, the model to be trained $\theta_i$ in institution $i$ is duplicated to $\theta_i^*$, preserving the model parameters from the previous institution for generating data used in the review process. The $*$ symbol is used to indicate all components related to the review process. A set of vectors for each class $x_c^{*}$ are initialized from $N(0,1)$ which are fed into the model $\theta_i^*$ to output $\hat{y}_c^{*}$, where $x$, $c$, and $\hat{y}$, refer to the input, classes, and output logit of the DL model, respectively. The $x_c^{*}$ are optimized with stochastic gradient descent (Adam \cite{RN169}) to maximize mean output logit for each data sample of each class $\hat{y}_c^{*}$ until it exceeds $\hat{y}_c^{D^{'}}$, the measured mean logit from the previous dataset. This process aimed to generate data samples that the model $\theta_i^*$ interprets similarly to the previous dataset. Once $\hat{y}_c^{*}$ exceeds $\hat{y}_c^{D^{'}}$ for all $x_c^{*}$, the $x_c^{*}$ and $\hat{y}_c^*$ pairs are saved as a generated dataset ($D^*$) for the knowledge distillation step, in which the logit $\hat{y}_c^*$ is used as a soft target the training model $\theta_i$ must predict. The mean logits $\hat{y}_c^D$ to real and generated data are measured after training is complete in the institution, to be used in knowledge extraction in the next institution. The number of samples generated is a hyperparameter and does not have to equal the number of samples from the previous dataset, because the loss function is weighted-averaged to balance the relative effectiveness of the generated data. Knowledge extraction may be performed once but was repeated every epoch for more stochasticity.

In the knowledge distillation step, the model is jointly trained with the real and generated data. The loss function is expressed by (\ref{eq:loss_review})-(\ref{eq:loss_total}). The knowledge distillation \cite{RN96,RN188} loss function $\mathcal{L}_{review}$ uses soft targets for each class $\hat{y}_c^*$ from the generated data $D^*$ and the output logits $\hat{y}_c$ of the model being trained $\theta_i$. T denotes the temperature of the softmax function \cite{RN96}. The task loss function $\mathcal{L}_{real}$ is measured with the current real dataset $D$. The $\mathcal{L}_{review}$ and $\mathcal{L}_{real}$ are jointly optimized with $\lambda_{review}$ weighing relative importance based on the relative number of samples. The $N_{review}$ and $N_{real}$ refers to the accumulated number of samples the model was trained on and the number of samples in the current dataset $D$, respectively.

\begin{equation}
\label{eq:loss_review}
\mathcal{L}_{review}=-T^2 \sum_{c=1}^{N_{classes}} \hat{y}_c^* \log \hat{y}_c
\end{equation}
\begin{equation}
\label{eq:lambda_review}
\lambda_{review}= \frac{N_{review}}{N_{real}+N_{review}}
\end{equation}
\begin{equation}
\label{eq:loss_total}
\mathcal{L}=(1-\lambda_{review})\mathcal{L}_{real}+\lambda_{review}\mathcal{L}_{review}
\end{equation}

\subsubsection{Evaluation}
The AUROC and Matthews correlation coefficient (MCC) \cite{RN207} metrics are reported in the main text, while four additional metrics are described in Supplementary Figures 10–15. First, the AUROC was computed from the target labels $y$ and model prediction probabilities $\hat{y}$. Subsequently, the binary model predictions were computed by arbitrarily thresholding the model prediction probabilities at the point where specificity (true negative rate) was larger than 0.75. MCC, accuracy, F1-score, precision, recall, and specificity were measured using the binary model predictions and the target labels, using (\ref{eq:metric_MCC})-(\ref{eq:metric_tnr}).
\begin{equation}
\label{eq:metric_MCC}
MCC=\frac{TP\times TN-FP \times FN}{\sqrt{(TP+FP)(TP+FN)(TN+FP)(TN+FN)}}
\end{equation}
\begin{equation}
\label{eq:metric_accuracy}
Accuracy=\frac{TP+TN}{TP+TN+FP+FN}
\end{equation}
\begin{equation}
\label{eq:metric_f1-score}
F1=\frac{2\times Precision \times Recall}{Precision+Recall}
\end{equation}
\begin{equation}
\label{eq:metric_precision}
Precision=\frac{TP}{TP+FP}
\end{equation}
\begin{equation}
\label{eq:metric_recall}
Recall=\frac{TP}{TP+FN}
\end{equation}
\begin{equation}
\label{eq:metric_tnr}
Specificity=\frac{TN}{TN+FP}
\end{equation}

where TP, TN, FP, and FN stand for true positive, true negative, false positive, and false negative, respectively. Each metric was measured using data from all institutions and data within each institution, which we refer to as global and local score. Higher global score implies the model is well suited for fine tuning for specific institutions, and higher local score implies the model is suitable for deployment at the institution.

Evaluation was performed in a privacy-preserving manner to keep test data at its source (Fig. \ref{fig:model_evaluation}). A model was distributed to each institution and made predictions on the test data. Since the target labels $y$ and the predicted probability values $\hat{y}$ did not contain information that could identify each patient, they were not privacy sensitive. Thus, $y$ and $\hat{y}$ for test data were aggregated to compute the global scores. The local scores were measured using local $y$ and $\hat{y}$.

\subsection{Experimental Setup}

\begin{figure*}[!t]
\centering
\subfloat[]{\includegraphics[width=0.52\textwidth]{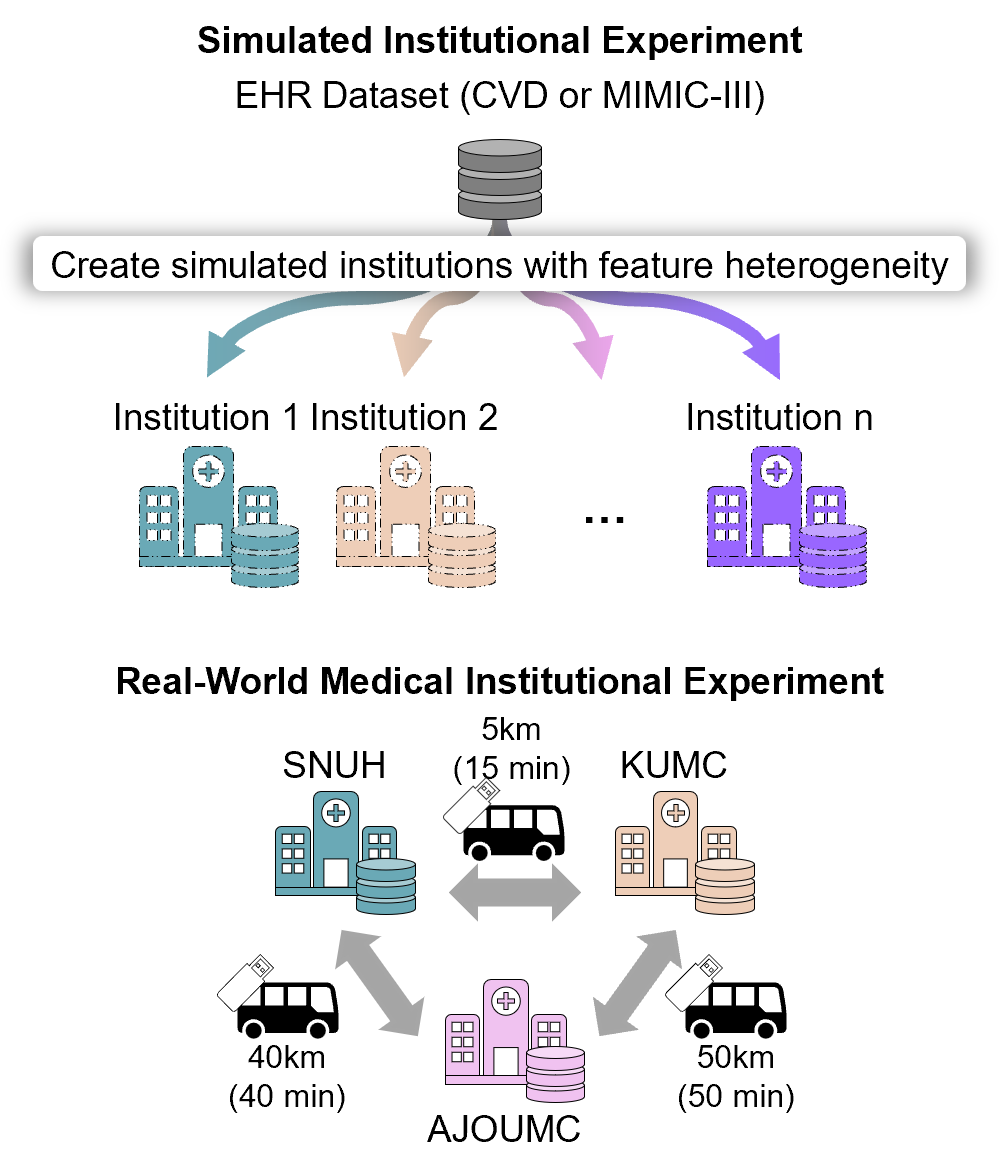}%
\label{fig:research_dataset}}
\hfil
\subfloat[]{\includegraphics[width=0.48\textwidth]{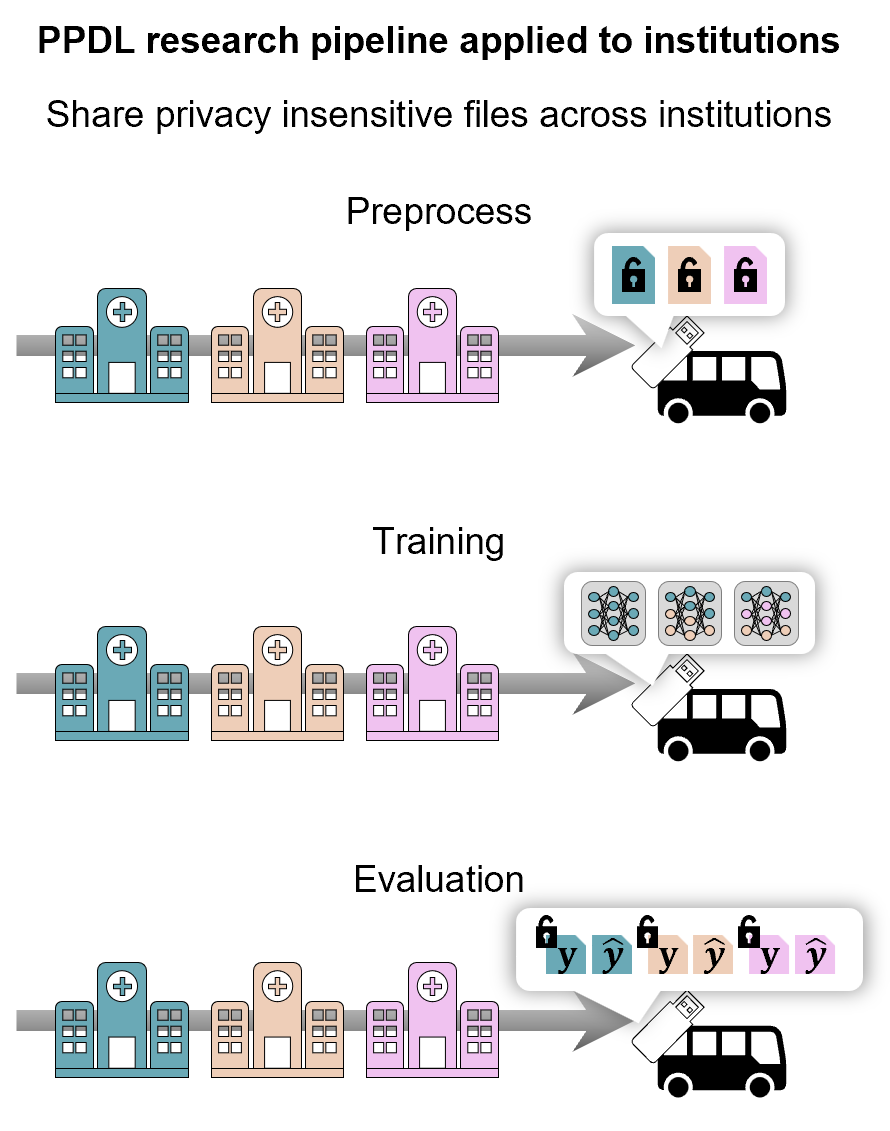}%
\label{fig:PPDL_to_institutions}}
\caption{Experiment overview. (a) Institutional data setting for experiments. (b) Privacy preserving deep learning (PPDL) research pipeline applied to institutions.}
\label{fig:research_overview}
\end{figure*}
Simulated institutional experiment was performed first to see the potential of RevL before applying it to real-world institutions, which required complex logistics such as transportation and stringent security measures (Fig. \ref{fig:research_overview}). To accommodate the need for multiple institutions in our experiments, we generated simulated institutions with feature heterogeneity across them, as described in Section \ref{sec:simulated_institutions}. All experiments were conducted using Python software.

\subsubsection{Simulated Institutional Experiment}
To evaluate the efficacy of RevL in real-world institutions, its performance must be compared with prevalent PPDL model development methods in medical practice. Therefore, in the simulated institutional experiment, RevL, TL, CDS \cite{RN52}, and LL were evaluated, serving as the positive control with continual learning, negative control that does not utilize continual learning, upper baseline, and lower baseline, respectively. CDS is an upper baseline in performance that could only be tested in simulation, because it trains a model using aggregated dataset which is not privacy-preserving. LL is a simple lower baseline in performance where a single model is trained within a local institution. TL serves as the negative control that RevL is expected to outperform. For RevL, TL, and LL, the training and validation were performed within each hypothetical institution, while CDS used aggregated training and validation sets for training and validation, respectively. For all preprocessing, training, and evaluation steps, a single node with a graphic processing unit (GPU) was used.

LL models used local statistics to standardize the continuous variables as they assume no collaboration between institutions. In contrast, RevL, TL, and CDS used global statistics from the training sets of all institutions (Methods \ref{sec:preprocess}), as these training algorithms aim to enable DL models to learn the global data distribution for the task.

All hyperparameters were empirically determined through initial CDS experiments and are as follows. After the hypothetical institutions were generated, data from each institution were divided by stratified random sampling into training, validation, and test sets with ratios of 0.7, 0.15, and 0.15, respectively. A fully connected neural network model with one hidden layer of 32 neurons and a single sigmoid output layer was used. A rectified linear unit (ReLU) followed by a dropout layer with a dropout probability of 50\% were applied to the hidden layers. The model parameters were optimized using the Adam \cite{RN169} optimizer with a learning rate of 1e-3. The batch size was set to 64 for MIMIC-III data and 256 for CVD prediction data. Binary cross-entropy was used as the loss function, and class weights were applied inverse proportionally to the number of samples for each class to prevent bias in prediction. The number of epochs was 100. Note that in RevL, TL, and LL, the model was trained for 100 epochs within each local institution, whereas in CDS, the single server model was trained for 100 epochs. Early stopping was applied by evaluating the area under the receiver operating curve (AUROC) score on the validation set. Validation was performed every 10 updates for MIMIC-III data and 20 steps for CVD prediction data, and the patience was set to 20. Each training algorithm was experimented with five different random seeds which affects DL model initialization.

The learning rate for the feature visualization in the RevL knowledge extraction step was 1e-2, and 512 samples were generated. The Adam optimizer was used in data generation, and gaussian noise was used as the initial data sample. A sigmoid function was applied to the binary variables as regularization so that the binary variables of the resulting data sample had values between 0 and 1. The temperature T from RevL knowledge distillation was set to 5.

\subsubsection{Real-World Institutional Experiment}
The experiment was conducted by manually visiting SNUH, KUMC, and AJOUMC for the preprocessing, training, and evaluation steps, because the installation of an automated software was restricted owing to their security policies (Fig. \ref{fig:research_overview}). The manual implementation without an automated software was possible as TL based PPDL approaches require minimal communication between institutions. For all preprocessing, training, and evaluation steps, a single node with a GPU was used in SNUH and AJOUMC while a single laptop with a central processing unit (CPU) without a GPU was used at KUMC. Privacy-insensitive files were exported after security inspections at each institution. 

RevL, TL, and LL were tested in the real-world multi-institutional experiment. The CDS method was not implemented because privacy-sensitive data could not be aggregated in the real-world experiment. The local statistics of continuous variables, model parameters, and prediction results were shared while keeping the original privacy-sensitive data secure at each institution.

Hyperparameters were set empirically through initial experiments using SNUH data. The data from each institution were divided by stratified random sampling into training, validation, and test sets with ratios of 0.7, 0.15, and 0.15, respectively. A fully connected neural network model with hidden layers of 128, 64, and 32 neurons was used. A ReLU followed by a dropout layer with a dropout probability of 50\% were applied to the hidden layers, and a sigmoid function was used in the output layer. The Adam optimizer was used to train the model; the batch size was 128, the learning rate 1e-3, the weight decay 1e-3, and the number of epochs 40. Binary cross-entropy was used as the loss function, and class weights were applied in the same way as in the simulation experiment. Early stopping was performed by evaluating the AUROC score on the validation set every 20 updates, and the patience was set to 20. Each training algorithm was experimented three times with different random seeds. The RevL hyperparameters were identical to those in the simulation experiment.

\section{Materials}

\subsection{Data for Simulated Institutional Experiments}

Domain incremental learning was experimented under PPDL restriction, a type of continual learning \cite{van2022three} to learn the same task on different datasets. Two binary classification EHR data were used in the simulation experiment. The first data was a 30-day mortality prediction of sepsis-3 patients \cite{RN189} from the Medical Information Mart for Intensive Care III database (MIMIC-III) \cite{RN190}. MIMIC-III is a publicly accessible EHR database from Beth Israel Deaconess Medical Center in Boston, Massachusetts, USA. This database consists of 58,976 admissions and 46,520 unique patients admitted to the intensive care unit between 2001 and 2012. Supplementary data provided by Nianzong et al. \cite{RN189} were refined for analysis. Rows with invalid age values were removed and feature columns were selected. There were 10 binary columns and 67 continuous columns, an input vector of size 77 in total. The refined data is available from the public database (https://doi.org/10.21227/qrhd-a469) \cite{RN208}.

The second type of EHR data was a cardiovascular disease (CVD) prediction for diabetes patients from SNUH. The study population was extracted from a common data model \cite{RN1} of SNUH. The task definition was to predict CVD occurrence one year after the patients were diagnosed with diabetes. The initial study population consisted of 52,718 patients who had been diagnosed with type 2 diabetes \cite{RN191} between 2004 to 2020. However, the patients who were diagnosed with CVD within one year or before diabetes diagnosis were removed since CVD could have been caused by other factors. The patients without any features other than age and sex were also removed. The final patients with CVD were defined as the outcome group, and the rest as the control group. The resulting study population consisted of 40,507 patients, with 3,478 and 37,029 patients in the outcome and control groups, respectively. There were two demographic variables (age, sex), 160 condition variables, 27 measurement variables, and 5 medication variables. The categorical variables were converted into binary columns of each category. There were 167 binary columns and 28 continuous columns, an input vector of size 195 in total. In the case of multiple records in continuous columns for the same patient, the last record was used as the representative value. This study was approved by the Institutional Review Board (IRB) of SNUH (No. 2210-042-1366).

\subsection{Creating Simulated Institutions with Feature Heterogeneity}
\label{sec:simulated_institutions}
\begin{table}
    \centering
    \begin{tabular}{ccrrrr}
         \hline
         Institutional setting & Institution name & Control & Case & N & Case ratio\\
         \hline
         \multirow{3}{*}{3} & Local 1 & 2,089 & 115 & 2,204 & 5.22\% \\
         & Local 2 & 1,211 & 219 & 1,430 & 15.31\% \\
         & Local 3 & 157 & 449 & 606 & 74.09\% \\
         \hline
         \multirow{4}{*}{4} & Local 1 & 1,517 & 113 & 1,630 & 6.93\% \\
         & Local 2 & 1,204 & 218 & 1,422 & 15.33\% \\
         & Local 3 & 634 & 305 & 939 & 32.48\% \\
         & Local 4 & 102 & 147 & 249 & 59.04\% \\
         \hline
         \multirow{5}{*}{5} & Local 1 & 1,210	&123	&1,333&	9.23\%  \\
         & Local 2 & 578	&285&	863&	33.02\% \\
         & Local 3 & 776	& 70	&846	&8.27\% \\
         & Local 4 & 508	& 178	&686	&25.95\%\\
         & Local 5 & 385	&127	&512	&24.80\% \\
         \hline
    \end{tabular}
    \caption{Summary of MIMIC-III 30-day mortality prediction of sepsis-3 patients data in simulated institutional experiment}
    \label{tab:summary_sim_mimic3}
\end{table}

\begin{table}
    \centering
    \begin{tabular}{ccrrrr}
         \hline
         Institutional setting & Institution name & Control & Case & N & Case ratio\\
         \hline
         \multirow{3}{*}{3} &  Local 1 & 16,817	&853	&17,670	&4.83\% \\
         & Local 2 & 14,360&	678	&15,038	&4.51\% \\
         & Local 3 & 5,852&	1,947&	7,799&24.96\% \\
         \hline
         \multirow{4}{*}{4} & Local 1 & 16,254&	764&	17,018	&4.49\% \\
         & Local 2 & 12,835	&103&	12,938	&0.80\%\\
         & Local 3 & 3,929	&1,947	&5,876	&33.13\% \\
         & Local 4 &4,011&	664	&4,675	&14.20\%\\
         \hline
         \multirow{5}{*}{5} & Local 1 & 16,254&	83	&16,337	&0.51\%  \\
         & Local 2 &9,512	&97&	9,609&	1.01\% \\
         & Local 3 & 6,171&	400&	6,571&	6.09\%\\
         & Local 4 & 3,662&	951&	4,613&	20.62\% \\
         & Local 5 & 1,430&	1,947&	3,377&	57.65\% \\
         \hline
    \end{tabular}
    \caption{Summary of cardiovascular disease prediction for diabetic patients data in simulated institutional experiment}
    \label{tab:summary_sim_CVD}
\end{table}

Domain-incremental learning experiment under PPDL restriction requires institutional datasets with feature heterogeneity to reproduce the forgetting problem. For the simulated institutional experiment, $n$ simulated institutions were created as follows. First, the data were divided into control and outcome groups. Second, the $2n$ Gaussians were fit using a Gaussian mixture model (GMM) \cite{RN206} for each group. The GMM fits a fixed number of Gaussians from floating point data, which can be used to calculate the probability of each data sample belonging to a particular Gaussian. Thus, only continuous columns were used in fitting the GMM. Third, the patients from the control and outcome groups were randomly assigned to $2n$ groups according to the density of Gaussians. Fourth, each control and outcome group were paired to form $n$ institutional datasets. There are $n!$ possible pairs, each representing a potential institutional setting. For each setting, $n$ local logistic regression models were fit on each institutional dataset. Fifth, the angles between every logistic regression model weight vector pairs were measured. There were $\binom{n}{2}$ angles as there are $n$ weight vectors from models. The mean angles demonstrate the data heterogeneity of the current institutional setting. Note that the weight vector is equivalent to the normal vector of the hyperplanes created by the logistic regression model. Finally, the institutional setting with the maximum mean angle was selected to maximize feature heterogeneity. Three numbers of hypothetical institutional settings were tested: three, four, and five. The hypothetical institutions were named “local 1,” “local 2,” …, and "local n" in descending order of data size (Table \ref{tab:summary_sim_mimic3}, \ref{tab:summary_sim_CVD}). The baseline characteristics for simulated institutions describing the feature columns and their statistics are presented in Supplementary Fig. 1–6.

\subsection{Data for Real-World Medical Institutional Experiment}
The real-world multi-institutional experiment was performed using the CVD prediction of diabetic patients data from SNUH, KUMC, and AJOUMC. The data were sequestered in remote medical institutions within South Korea. The task definition was identical to that in the simulation experiment, but the variables of the data were modified into common variables across three medical institutions. Patients diagnosed with type 2 diabetes \cite{RN191} between 2004 to 2020 were included in the data from each institution. The numbers of patients in the outcome and control groups are listed in TABLE \ref{table:summary_real_institutions}. There were two demographic variables (age, sex), 10 condition variables, 25 measurement variables, and 19 medication variables which were 30 binary columns and 26 continuous columns in total. The baseline characteristics of institutional EHR data describing the feature columns and their statistics are presented in Supplementary Fig. 7. This study was approved by the IRB of SNUH (2210-042-1366), KUMC (2021AN0236), and AJOUMC (AJIRB-MED-MDB-21-193).

\begin{table}[!t]
\caption{Summary of cardiovascular disease (CVD) prediction data in real-world medical institutional experiment. Abbreviations: Seoul National University Hospital (SNUH), Korea University Medical Center (KUMC), and Ajou Medical Center (AJOUMC).}
\centering
\begin{tabular}{ccccc}
\hline
Institution name & Control & Case & N & Case ratio\\
\hline
SNUH & 38,585 & 3,381 & 41,996 & 8.06\% \\
\hline
KUMC & 33,878 & 3,192 & 37,070 & 8.61\% \\
\hline
AJOUMC & 25,067 & 2,375 & 27,442 & 8.65\% \\
\hline
\end{tabular}
\label{table:summary_real_institutions}
\end{table}

\section{Results}
\subsection{Inspection of Simulated Institutions and Review Learning}
The simulated institutional study was first performed to develop RevL. Hypothetical institutions with heterogeneous linear features were generated to simulate the catastrophic forgetting problem under the PPDL setting. An example of heterogeneous data distribution across institutions is presented in Supplementary Fig. 8. The RevL generates samples that represents what the model has learned using feature visualization \cite{RN155}, to retain knowledge by jointly training the model with the generated and real data. To ensure that relevant knowledge was extracted from the model, the generated samples were compared with the original data distribution (Supplementary Fig. 9). The t-distributed stochastic neighbor embedding was used to nonlinearly map and compare the real and generated data heuristically, as the generated data may contain nonlinear features that are hard to be quantified. The real and generated data were placed in similar directions for each of the control and outcome groups, indicating that relevant knowledge could be extracted through the knowledge extraction process of RevL.

\subsection{Simulated Institutional Experiment}

\begin{table*}[!t]
\caption{
Minimum and maximum global AUROC scores across institutional models. The scores are mean values across random repetitions. Highest scores among minimum AUROC are presented in bold font. The suffixes “asc” and “desc” indicate the ascending and descending order of institutional transfers according to the data size. Abbreviations: area under the receiver operating curve (AUROC), review learning (RevL), transfer learning (TL), local learning (LL), cardiovascular disease (CVD).}
\centering

\resizebox{\textwidth}{!}{
\begin{tabular}{c|ccccc|ccccc}

\hline
\multirow{2}{*}{\begin{tabular}{c} Simulation data, \\ \# of institutions
\end{tabular}}
& \multicolumn{5}{c|}{Minimum   AUROC}            & \multicolumn{5}{c}{Maximum   AUROC}             \\ \cline{2-11} 
                                      & RevL\_asc & TL\_asc & RevL\_desc & TL\_desc & LL    & RevL\_asc & TL\_asc & RevL\_desc & TL\_desc & LL    \\ \hline
MIMIC,   3        & \textbf{0.650}   & 0.612   & 0.626    & 0.587    & 0.578 & 0.816   & 0.829   & 0.792    & 0.763    & 0.746 \\ \hline
MIMIC,   4         & 0.541   & 0.528   & \textbf{0.675} & 0.591    & 0.531 & 0.773   & 0.753   & 0.797    & 0.783    & 0.730 \\ \hline
MIMIC,   5         & 0.656   & 0.578   & \textbf{0.675}    & 0.543    & 0.536 & 0.801   & 0.794   & 0.761    & 0.757    & 0.736 \\ \hline
CVD,   3              & 0.706   & 0.640   & \textbf{0.713}    & 0.625    & 0.540 & 0.863   & 0.855   & 0.849    & 0.849    & 0.840 \\ \hline
CVD,   4               & \textbf{0.717}   & 0.648   & 0.578    & 0.578    & 0.563 & 0.846   & 0.846   & 0.837    & 0.848    & 0.841 \\ \hline
CVD,   5               & \textbf{0.711}   & 0.667   & 0.570    & 0.570    & 0.575 & 0.842   & 0.833   & 0.827    & 0.819    & 0.828 \\ \hline
\end{tabular}
}

\label{table:simulation_performance}
\end{table*}

\begin{figure*}[!t]
\centering
\includegraphics[width=\textwidth]{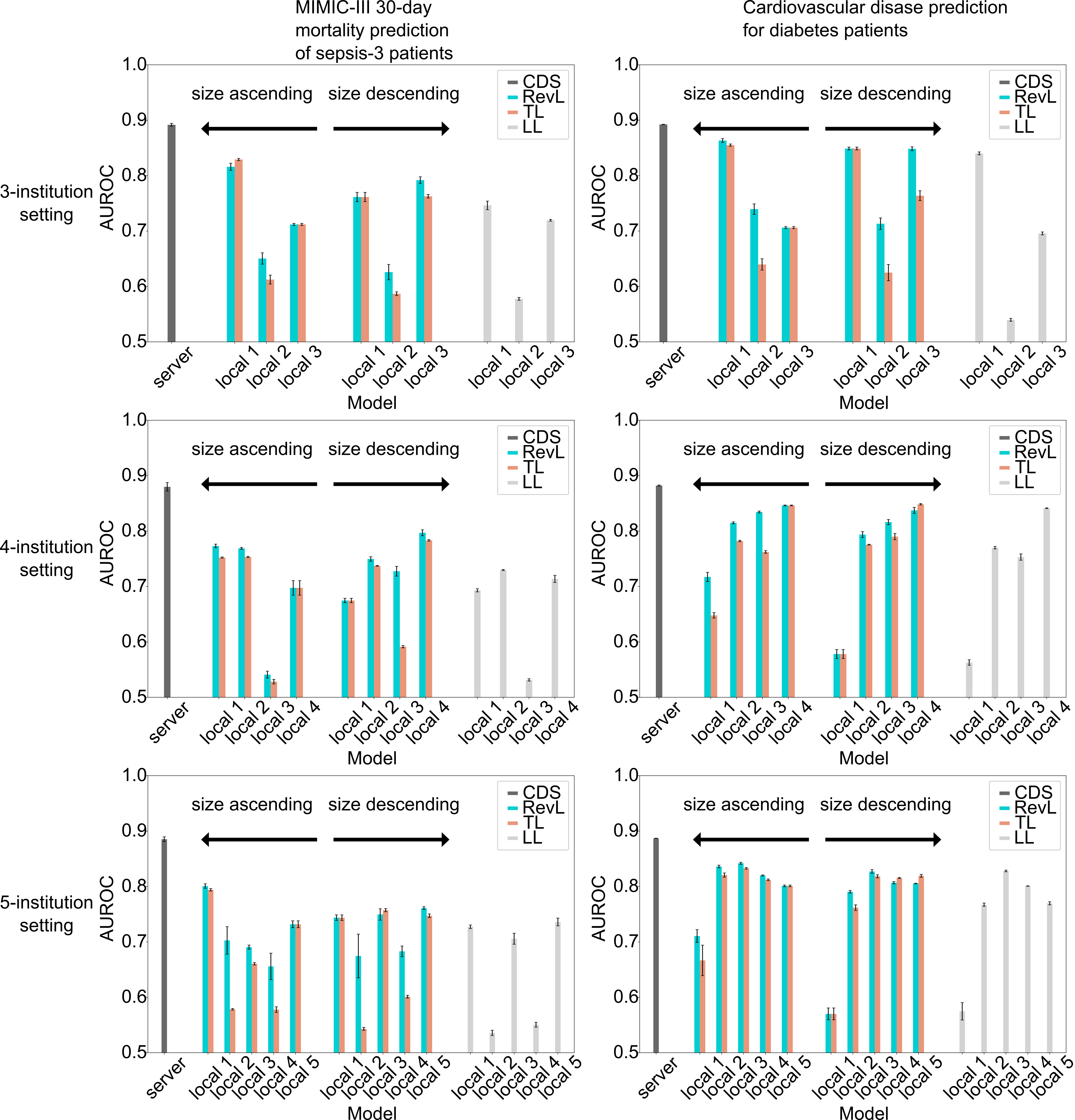}
\caption{Global scores of simulated institutional experiment measured in AUROC. Abbreviations: Central data sharing (CDS), Review learning (RevL), Transfer learning (TL), Local learning (LL). The error bars indicate the standard error.}
\label{fig:simulation_global_score}
\end{figure*}

\begin{figure*}[!t]
\centering
\includegraphics[height=0.85\textheight]{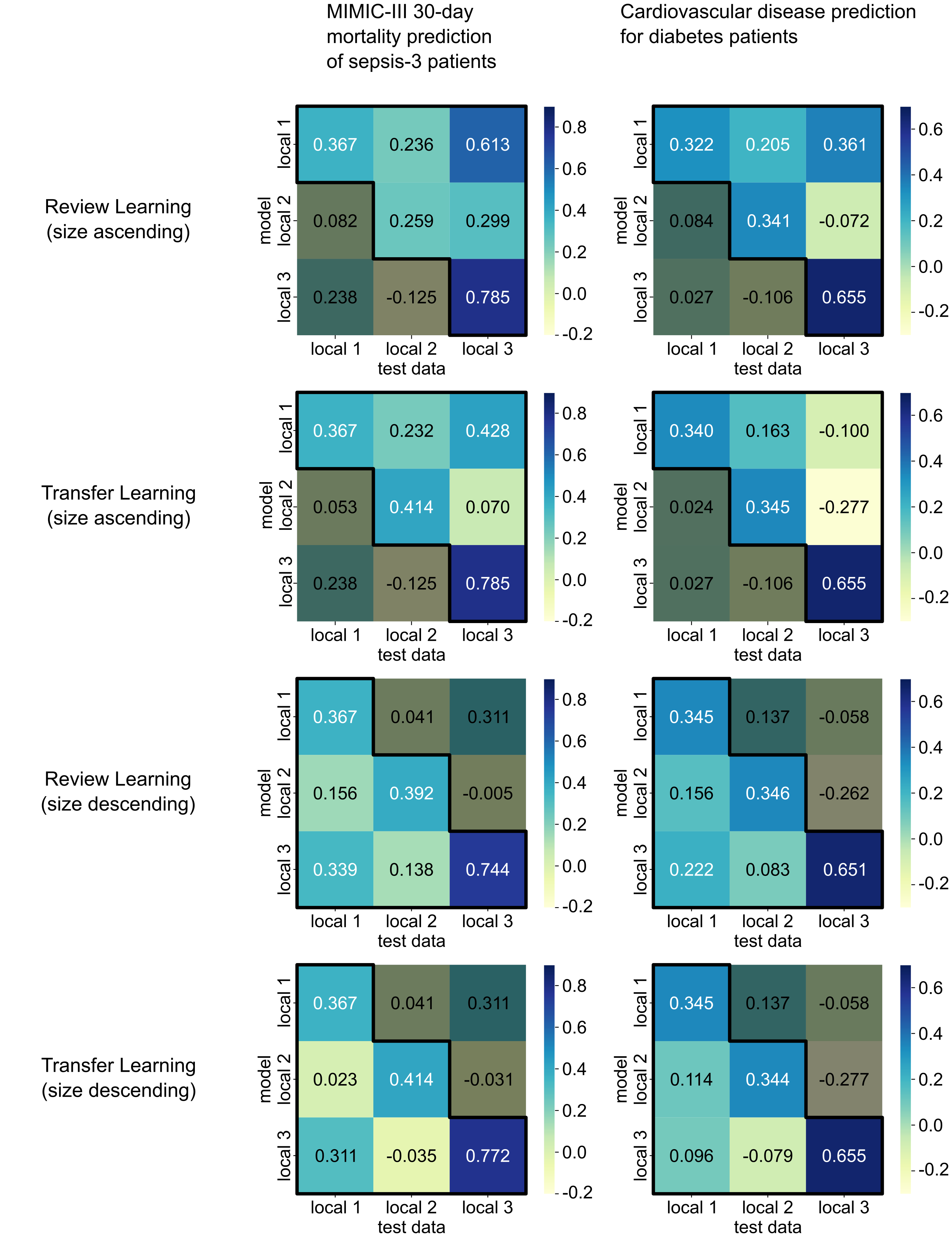}
\caption{Local scores of simulation experiment for RevL and TL, three-institutional setting. The mean of Matthews correlation coefficient (MCC) was measured. The performance scores on test data from unvisited institutions are shadowed.}
\label{fig:simulation_local_score}
\end{figure*}

The RevL, TL, CDS, and LL models were evaluated on two EHR binary classification data under three numbers of simulated institutions, resulting in six simulated institutional settings. The RevL and TL models were trained by visiting the institutions in ascending and descending order of data size, referred to as RevL\_asc, RevL\_desc, TL\_asc, and TL\_desc, respectively.

First, the global scores measured in the AUROC are presented in TABLE \ref{table:simulation_performance} and Fig. \ref{fig:simulation_global_score}. The TL models suffered from the catastrophic forgetting problem because the global score fluctuated significantly across institutional model transfers. However, the RevL models suffered less, shown from the higher minimum global score compared to TL models (TABLE \ref{table:simulation_performance}). This improvement is also visible in Fig. \ref{fig:simulation_global_score}, where the global scores of RevL models consistently outperform those of the TL models across all simulated institutional settings.

While RevL models consistently improved minimum global scores, the maximum global scores did not show consistent improvement (TABLE \ref{table:simulation_performance}). Thus, the continual learning effect of RevL prevented the models from forgetting previous knowledge but did not significantly enhance the maximum global scores.

The global scores of CDS model were the highest in the six settings, which is expected since the CDS model trains on aggregated data. LL models always underperformed compared to RevL and TL models except for the five-institutional setting from diabetes dataset, where the performance of LL model was comparable to those of RevL and TL models. Note that the initial models from RevL and TL were identical. Also note that the initial models from RevL and TL were different from their corresponding LL models because PPDL models used global statistics in preprocessing whereas LL models used local statistics because no collaboration is assumed between institutions.

The local scores measured in MCC are presented in Fig. \ref{fig:simulation_local_score}; for conciseness, only the three-institutional setting is shown. The catastrophic forgetting problem from TL models is evident from the local scores, as their performance on test data from previous institutions drops drastically, as seen in the off-diagonal entries of Fig. \ref{fig:simulation_local_score}. TL models achieved high local scores on recently trained data, corresponding to the diagonal entries in Fig. \ref{fig:simulation_local_score}. In contrast, RevL models consistently maintained higher local scores on test data from previous institutions compared to TL models, even when their global scores were similar (e.g., local 1 model from RevL\_asc from both simulation data). This indicates that the knowledge required for global and local classification is not identical, with RevL models retaining the knowledge for local classification while TL models lost it. Similar local score patterns were observed in other institutional settings, and all scores from the simulated institutional experiments are described in Supplementary Fig. 10-13.

\subsection{Real-World Institutional Experiment}
\begin{figure*}[!t]
\centering
\subfloat[]{\includegraphics[width=0.7\textwidth]{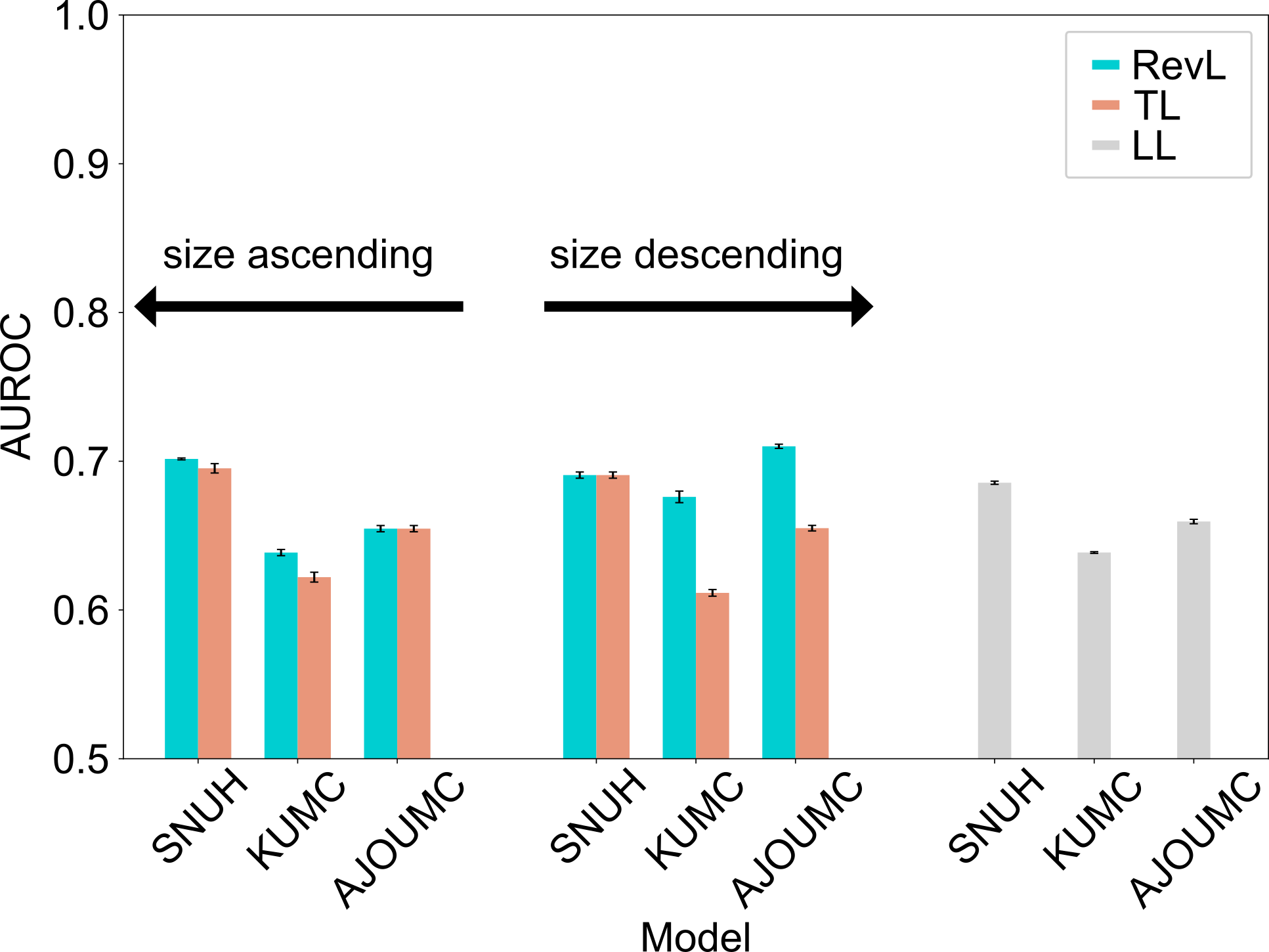}%
\label{fig:real_global_score}}
\hspace{0.1\textwidth}
\subfloat[]{\includegraphics[width=0.7\textwidth]{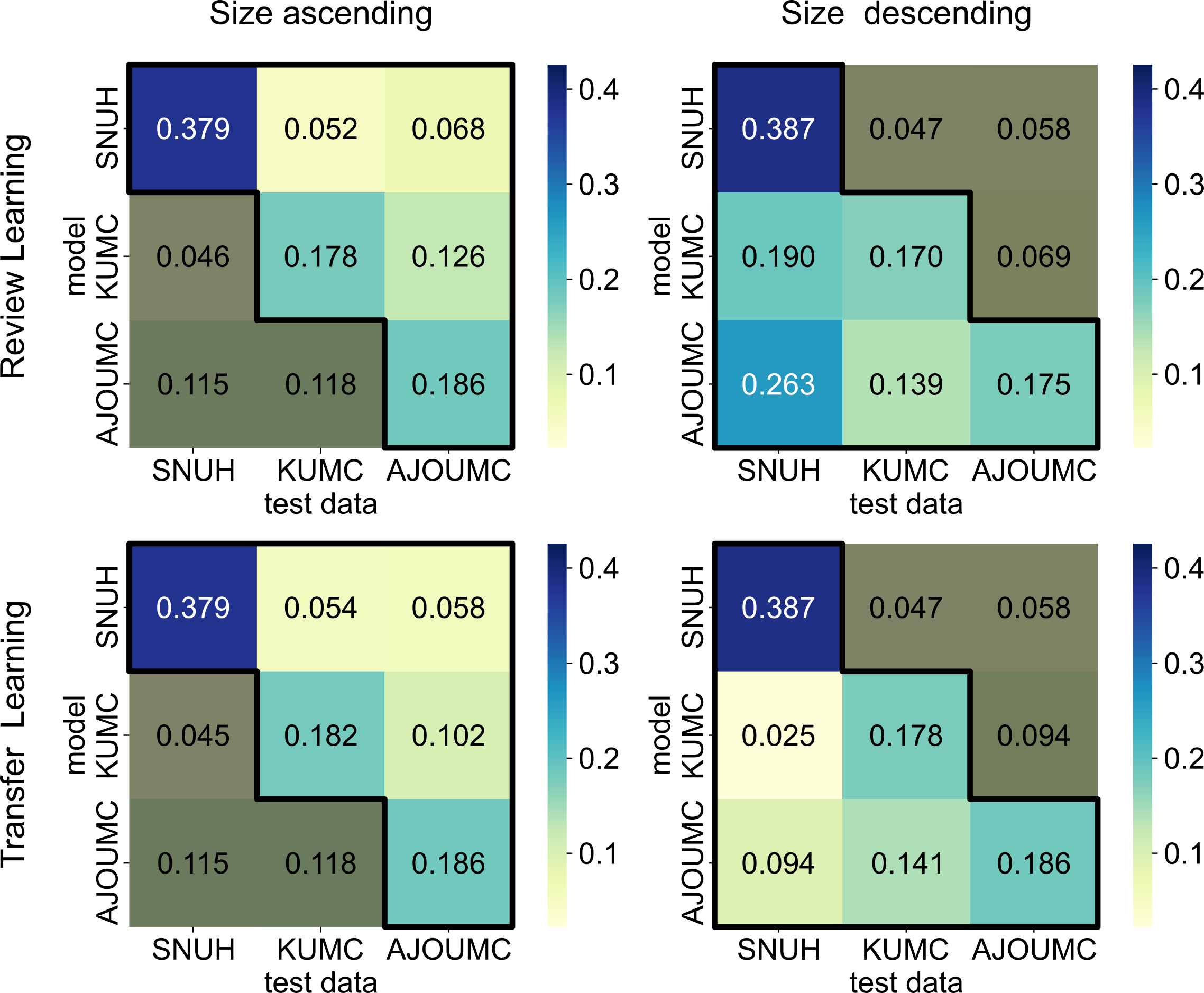}%
\label{fig:real_local_score}}
\caption{Performance scores of real-world multi-institutional experiment. (a) Global score measured in AUROC. The error bars indicate the standard error. (b) Local score. MCC is presented. Abbreviations: Seoul National University Hospital (SNUH), Korea University Medical Center (KUMC), Ajou Medical Center (AJOUMC).}
\end{figure*}

After identifying the potential of RevL in simulated institutional experiments, its performance was further validated in a real-world institutional PPDL experiment using data from SNUH, KUMC, and AJOUMC. The data size decreased sequentially from SNUH to KUMC and AJOUMC (TABLE \ref{table:summary_real_institutions}). The RevL, TL, and LL models were evaluated, while the CDS model could not be trained because data could not be aggregated in the real-world experiment.

The feature distribution across the three real institutions was heterogeneous, where the forgetting problem can be identified from the fluctuating global score of TL models (Fig. \ref{fig:real_global_score}) and the low off-diagonal values of TL local scores (Fig. \ref{fig:real_local_score}). 

RevL outperformed TL models in the real-world experiment as well. All models from RevL\_asc and RevL\_desc achieved higher global scores than their corresponding TL models. The most significant performance improvement was observed in the final model at AJOUMC, where the RevL\_desc model achieved a global AUROC of 0.710±0.002, compared to 0.655±0.003 for the TL\_desc model (p-value=0.05, Mann-Whitney U Test). A notable improvement in local scores was also seen in RevL\_desc. The performance of the model from the last institution (AJOUMC) on the test set from the first institution (SNUH) had a local MCC score of 0.263±0.013, significantly higher than the TL\_desc model's score of 0.094±0.014 (p-value=0.05, Mann-Whitney U Test). This local score improvement can be identified by comparing the off-diagonal entries of RevL\_desc and TL\_desc models (Fig. \ref{fig:real_local_score}). The RevL\_asc models had slightly higher global AUROC scores than the TL\_asc models, but the local scores of MCC did not improve drastically. All scores are described in Supplementary Fig. 14-15.

\section{Discussion}
Catastrophic forgetting prevents the deep learning model from acquiring knowledge sequentially, restricting the model from learning diverse features across various data. PPDL is one major field of TL application with its advantage of minimal asynchronous communication compared to federated learning. TL is particularly beneficial for medical institutions that restrict file sharing or lack the infrastructure to support automated federated learning platforms. However, catastrophic forgetting is a major obstacle since the purpose of PPDL is to train a model to learn knowledge from diverse privacy-sensitive data without sharing them. The knowledge across data from diverse sources is likely to be heterogeneous \cite{RN36,RN37}, and continual learning needs to be applied to TL to improve performance.

RevL, a new replay-based continual learning method for EHR data, was validated in this study through PPDL experiments with simulated and real-world institutions. RevL generates data within the main EHR classification model using feature visualization \cite{RN155}, eliminating the need for a separate generator and thereby reducing training complexity and the overhead of institutional transfers. The results showed that RevL outperformed TL in both global and local scores on the test sets from previous institutions, which demonstrates the continual learning effect of RevL, while remaining within score range between CDS and LL. As data quality and classification-relevant features are heterogenous across institutions, TL and LL models tend to develop poor global decision boundaries. However, PPDL continual learning methods like RevL could enhance performance by leveraging knowledge from other institutions, as identified by the higher minimum global scores achieved by RevL models across institutions (Table. \ref{table:simulation_performance}). Unlike previous works on brain-inspired generative replay, RevL did not require a separate generator, which removed the burden of training a generator and the cost of transferring it. All experiments were performed under PPDL research pipeline, ensuring data were securely sequestered in each institution.

The key to RevL is its review process, and the details of reviewing must be thoroughly analyzed so that the model could effectively review previously learned knowledge. One critical aspect of reviewing is the hyperparameters and regularizations. The two steps of the review process, feature visualization \cite{RN155} and knowledge distillation \cite{RN188}, may vary across data with different modalities, such as image or signals. In the case of generative replay using a generator, the generator learns the relevant data distribution, which includes regularization. However, regularizations are necessary for feature visualization because the data samples may be over-optimized without them. Regularizations suitable for certain data modalities, such as random jitter or frequency regularizations can be applied according to the task definition of the input data. In PPDL application, it is safe to compare the generated and real data samples before each institutional transfer because the real samples from previous institutions cannot be accessed for comparison after institutional transfers.

During knowledge extraction, the generated samples do not have to be realistic as long as the important representative features for the task can be identified. Unlike the objective of generative models which is to create realistic data, RevL needs samples to review knowledge for task solving such as classification. Therefore, rather unrealistic generated samples may be sufficient to retain the knowledge of the model. For instance, the binary columns of generated samples in our study contained continuous values between 0 and 1. Although the binary columns of real data only included the values of 0 or 1, the unrealistic generated samples were sufficient to retain the previous knowledge because in the end the deep learning model interprets all inputs as continuous variables. This is also evident from the real and generated data being positioned in similar directions for each of the control and outcome groups (Supplementary Fig. 9), suggesting that relevant knowledge can be extracted even when the generated samples are not strictly realistic.

RevL could be improved from several aspects. First, RevL could be compared with other continual learning methods. This study focused on validating RevL's efficacy in a real-world PPDL setting using EHR data, but its performance against other continual learning algorithms remains unknown. There were practical limitations in implementing other continual learning methods across real-world institutions, including limited computational resources (lack of GPU nodes in some institutions) and limited time access to such resources, and the need for extensive PPDL pipeline adjustments for multiple algorithms. Fortunately, replay-based continual learning methods often outperform other continual learning approaches including domain incremental learning \cite{van2022three}, indirectly underscoring RevL's continual learning potential. Nevertheless, incorporating various types of continual learning algorithms, such as elastic weight consolidation \cite{RN26} and synaptic intelligence \cite{RN185} (regularization approaches), dynamically expandable network \cite{yoon2017lifelong} (dynamic architectures), and brain-inspired replay using generative models \cite{RN96} (memory replay) would enhance the understanding and efficacy of PPDL research in healthcare. Another potential is extending to different data modalities. RevL was validated using EHR data and different data-specific adjustments needs investigation such as regularization and hyperparameter tuning. To ensure sensitive data remains unretrievable against privacy attacks \cite{fredrikson2015model,zhang2020secret}, safeguarding techniques \cite{abadi2016deep} may be added as well. Hyperparameter and model architecture search under PPDL setting also needs investigation for finer tuning. The impact of institutional visit order also needs further investigation, which is analogous to the task ordering in continual learning, an active area of research with limited understanding \cite{pmlr-v202-lin23f, bell2022effect, hacohen2024forgetting}. With limited literature and a focus on real-world validation, this study only considered data size when determining institutional visit order. Finally, the performance of real multi-institutional experimental data needs improvement. RevL consistently outperformed TL with statistical significance in all institutional settings including the real-world experiment, validating its real-world efficacy in PPDL settings with EHR data. However, the real institutional RevL model with the best global score was an AUROC of 0.710±0.002 and an MCC of 0.182±0.002, which was not particularly large. Improvements could involve enhanced preprocessing to better encode EHR data for classification, comparing with task-specific model literature \cite{kavitha2024healthcare}, and detailed data curation at each institution to ensure the train and the test data are not poisoned. A higher task performance score from the real experiment could more clearly confirm the efficacy of RevL in PPDL settings with EHR data.

\section{Conclusion}
In this work, we demonstrated the efficacy of continual learning using EHR data in a real PPDL setting. First, we introduced a comprehensive PPDL research pipeline for DL, encompassing preprocessing, training, and evaluation, through sharing privacy insensitive data while securing privacy sensitive data. Next, we presented a continual learning method named review learning, a generator-free replay-based continual learning algorithm designed for EHR data. Finally, we validated the effectiveness of review learning using our proposed PPDL pipeline through two simulated institutional experiments and one real-world institutional experiment involving 3 medical institutions. The results from both simulated and real-world institutional experiments show that review learning effectively prevented performance loss (catastrophic forgetting) in both global and local scores against data from previously trained institutions. Our work establishes a pipeline for PPDL research and highlights the practical potential of continual learning to train a generalizable DL model with real medical institutions under a secure PPDL framework. Future work could enrich PPDL continual learning research by comparing alternative continual learning algorithms, exploring different data modalities, analyzing the effect of institutional visit order, and enhancing real-world performance through fine-tuning.

\section*{Acknowledgements}
This research was supported by the MHW (Ministry of Health and Welfare) of Korea after the project named DisTIL (Development of Privacy-Reinforcing Distributed Transfer-Iterative Learning Algorithm; Grant number: HI19C0842), supervised by the KHIDI (Korea Health Industry Development Institute). This work was also supported by the Bio Industrial Strategic Technology Development Program (20003883) funded By the Ministry of Trade, Industry \& Energy (MOTIE, Korea). During the preparation of this work the authors used chatGPT (https://chat.openai.com/) only to edit the English sentence structure for better delivery. Corresponding author: Kwangsoo Kim (e-mail: kim.snuh@gmail.com).
 \bibliographystyle{elsarticle-num}
 \bibliography{ReviewLearning}





\end{document}